\documentclass{ecai} 
\usepackage{latexsym}
\usepackage{amssymb}
\usepackage{amsmath}
\usepackage{amsthm}
\usepackage{booktabs}
\usepackage{enumitem}
\usepackage{graphicx}
\usepackage{color}
\usepackage{multirow} 

\newtheorem{definition}{Definition}

\newcommand{\BibTeX}{B\kern-.05em{\sc i\kern-.025em b}\kern-.08em\TeX}

\begin{document}

%%%%%%%%%%%%%%%%%%%%%%%%%%%%%%%%%%%%%%%%%%%%%%%%%%%%%%%%%%%%%%%%%%%%%%%%

\begin{frontmatter}

%%% Use this command to specify your submission number.
%%% In doubleblind mode, it will be printed on the first page.

% \paperid{2356} 

%%% Use this command to specify the title of your paper.

\title{Multivariate Time Series Forecasting with Hybrid Euclidean-SPD Manifold Graph Neural Networks}

%%% Use this combinations of commands to specify all authors of your 
%%% paper. Use \fnms{} and \snm{} to indicate everyone's first names 
%%% and surname. This will help the publisher with indexing the 
%%% proceedings. Please use a reasonable approximation in case your 
%%% name does not neatly split into "first names" and "surname".
%%% Specifying your ORCID digital identifier is optional. 
%%% Use the \thanks{} com、mand to indicate one or more corresponding 
%%% authors and their email address(es). If so desired, you can specify
%%% author contributions using the \footnote{} command.

\author[A]{\fnms{Yong}~\snm{Fang}}
\author[B]{\fnms{Na}~\snm{Li}}
\author[C]{\fnms{Hangguan}~\snm{Shan}\thanks{Corresponding Author. Email: hshan@zju.edu.cn.}} 
\author[D]{\fnms{Eryun}~\snm{Liu}}
\author[E]{\fnms{Xinyu}~\snm{Li}}
\author[F]{\fnms{Wei}~\snm{Ni}}
\author[G]{\fnms{Er-Ping}~\snm{Li}}

\address[A,B,C,D,G]{Zhejiang University}
\address[E]{Huazhong University of Science and Technology}
\address[F]{The University of New South Wales}
%%% Use this environment to include an abstract of your paper.

\begin{abstract}
Multivariate Time Series (MTS) forecasting plays a vital role in various real-world applications, such as traffic management and predictive maintenance. Existing approaches typically model MTS data in either Euclidean or Riemannian space, limiting their ability to capture the diverse geometric structures and complex Spatio-Temporal (ST) dependencies inherent in real-world data. To overcome this limitation, we propose the \textbf{H}ybird \textbf{S}ymmetric Positive-Definite \textbf{M}anifold \textbf{G}raph \textbf{N}eural \textbf{N}etwork (HSMGNN), a novel graph neural network-based model that captures data geometry within a hybrid Euclidean–Riemannian framework. To the best of our knowledge, this is the first work to leverage hybrid geometric representations for MTS forecasting, enabling expressive and comprehensive modeling of geometric properties. Specifically, we introduce a Submanifold-Cross-Segment (SCS) embedding to project input MTS into both Euclidean and Riemannian spaces, thereby capturing ST variations across distinct geometric domains. To alleviate the high computational cost of Riemannian distance, we further design an Adaptive-Distance-Bank (ADB) layer with a trainable memory mechanism. Finally, a Fusion Graph Convolutional Network (FGCN) is devised to integrate features from the dual spaces via a learnable fusion operator for accurate prediction. Experiments on three benchmark datasets demonstrate that HSMGNN achieves up to 13.8\% improvement over state-of-the-art baselines in forecasting accuracy.
\end{abstract}

\end{frontmatter}

%%%%%%%%%%%%%%%%%%%%%%%%%%%%%%%%%%%%%%%%%%%%%%%%%%%%%%%%%%%%%%%%%%%%%%%%

\section{Introduction}

Multivariate Time Series (MTS) consists of multiple interdependent numerical variables evolving over time, crucial for modeling complex systems where variables interact across time and space, such as industrial equipment health monitoring, climate modeling, and healthcare diagnostics. In machinery Remaining Useful Life (RUL) prediction~\cite{FCSTGNN}, MTS forecasting enables proactive maintenance to prevent failures. However, the unique characteristics of MTS challenge traditional forecasting methods. Firstly, MTS exhibits dual dependencies: temporal dependency across timestamps and spatial dependency among variables at each timestamp. Secondly, existing Deep Learning (DL) methods typically map features into either Euclidean or Riemannian space. These complex Spatio-Temporal (ST) interactions demand modeling frameworks that capture both the geometric distributions and dependencies inherent in MTS.\footnote{Published in ECAI 2025.}

In recent works, researchers have explored temporal dependencies using various DL approaches with temporal encoders~\cite{jiang2022sequential,Li2023MTSMixersMT, wu2023timesnet}. However, these methods may be suboptimal when interdependencies exist among data points, as they overlook dynamic spatial relationships. 
To address this, Graph Neural Network (GNN) methods~\cite{chen2023multi,chen2024biased,TDSTGL2024,FCSTGNN,wu2020connecting,FourierGNN,zhou2023detecting,AnomalyDetection_ZQH} have been proposed to capture ST dependencies in MTS, constructing ST graphs by building separate graphs at each timestamp, with predefined or learnable structures for spatial dependencies and temporal encoders for temporal patterns. GNN-based approaches generally outperform those relying solely on temporal encoders~\cite{wang2024klinkllms}. 
Nevertheless, modeling spatial and temporal aspects independently often overlooks their intricate interactions. Moreover, existing GNNs tend to ignore asymmetric correlations between sensors at different timestamps~\cite{chen2024biased,FCSTGNN}, leading to insufficient modeling of dynamic node relationships in evolving adjacency matrices. Finally, most methods operate in a single (typically Euclidean) geometric space, neglecting diverse geometric structures critical for accurate forecasting.

% spd manifold
\begin{figure}[htb]
\centering
\includegraphics[width=0.75\columnwidth]{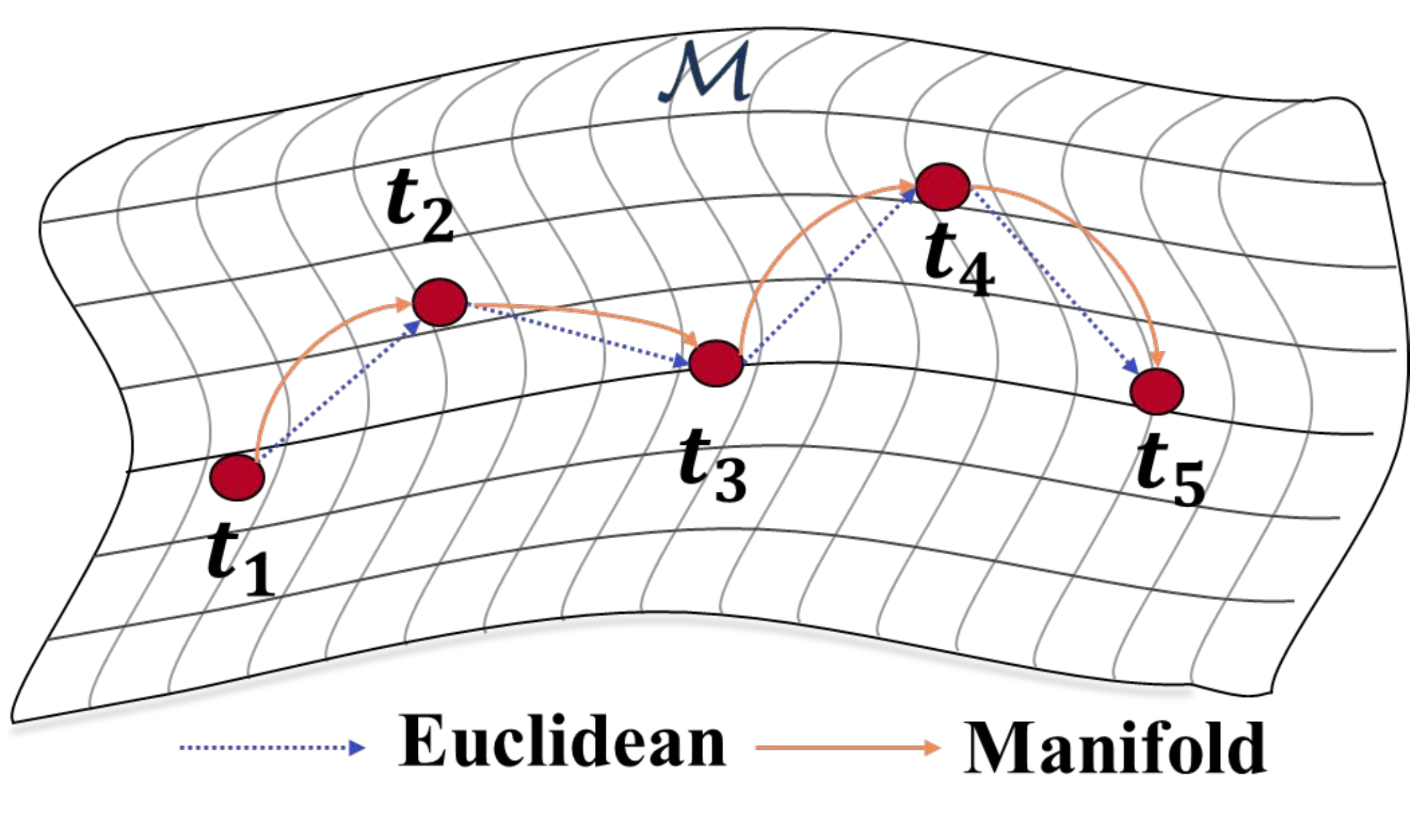} % Reduce the figure size so that it is slightly narrower than the column. Don't use precise values for figure width. This setup will avoid overfull boxes.
\caption{
% SPD manifold embedding, where $\mathrm{cov}\left(\mathbf{x}\right) = \mathbf{x}\mathbf{x}^{\text{T}}$ denotes the unnormalized covariance matrix, assuming that $\mathbf{x}$ has zero mean and represents a single sample.
Illustration of the difference between MTS representations in Euclidean and manifold spaces. 
% $\mathcal{S}_{++}^n$ denotes the $n \times n$ SPD manifold.
% Each MTS is embedded into either a Euclidean or a manifold space, where each red point represents the data at a specific time step. Arrows between consecutive points indicate temporal evolution, with their directions capturing the progression over time and their lengths reflecting the magnitude of change between time steps.
 }
\label{sub_ssmd}
\end{figure}

Fig. \ref{sub_ssmd} provides an example of an MTS evolving from time $t_1$ to time $t_5$ regarding both geometric structures. According to the manifold hypothesis~\cite{tenenbaum2000global}, high-dimensional data often lie on a low-dimensional manifold embedded in the ambient space. When ST dependencies are modeled using Euclidean techniques such as self-attention (indicated by blue arrows), the intrinsic geometric structure of the data lying on a low-dimensional manifold $\mathcal{M}$ cannot be faithfully preserved. In contrast, modeling transitions using Riemannian manifold metrics, indicated by orange arrows and measured via geodesic distances, captures the temporal evolution by respecting the data geometry. 
% This observation underscores the necessity of developing a hybrid Euclidean–SPD manifold framework for more accurate and geometry-aware time series forecasting.

Motivated by the above observation and the recent success of Geometric Learning combined with DL techniques, several methods have been proposed for MTS forecasting on Riemannian manifolds, particularly the \textit{Symmetric Positive-Definite (SPD)} manifold~\cite{katsman2024Riemannian,meilua2024manifold, nazari2023geometric}.
Such methods can naturally extend to non-Euclidean data structures like graphs and manifolds, making them well-suited for forecasting. Among these approaches, the SPD manifold has been actively studied due to its ability to learn beneficial statistical representation. As the first Riemannian network for non-linear matrix learning on SPD manifolds, SPDNet~\cite{SPDnet} uses the BiMap layer, ReEig layer, LogEig layer, and classical Euclidean network for visual classification tasks. Thereafter, many studies~\cite{huang2022Riemannian,islam2023revealing,Suh_Kim_2021_EEG_BANK,zhang2023STEEG} have laid the foundation for related fields, such as electroencephalogram decoding and nonlinear regression. These SPD-based forecasting methods introduce an architecture that integrates SPD manifold-based modules to extract ST information from primary signals. However, these techniques depend on computationally expensive operations of the Log-Euclidean metric and struggle to encode diverse geometric relationships into a single metric, limiting their ability to capture complex topological patterns under computational constraints.

In this paper, we propose \textbf{H}ybird \textbf{S}ymmetric Positive-Definite \textbf{M}anifold \textbf{G}raph \textbf{N}eural \textbf{N}etwork (HSMGNN) to capture both geometric distributions and dependencies inherent in MTS. Our approach includes three key aspects.
% enhances both the accuracy performance and the generalization capability of MTS forecasting across diverse spatiotemporal domains, 
First, we propose Submanifold-Cross-Segment (SCS) embedding to address the limitations of conventional single-metric-based models in capturing the complex geometric structure. In light of Fig. \ref{sub_ssmd}, SCS embeds Euclidean-derived representations into SPD submanifolds to capture geometric variations across both ST dimensions.
% This dual-space embedding module is explicitly designed to capture geometric variations across both ST dimensions, enabling more expressive and structure-aware feature representations for downstream forecasting.
% Resulting in smaller cross Riemannian patterns from conventional Euclidean patterns. Then SCS captures geometric representations 
Moreover, we introduce an Adaptive-Distance-Bank (ADB) layer to refine manifold representations with low computational cost. ADB encodes the adaptive weighting of signal embeddings within the SPD manifold across multiple submanifolds via a novel Nonlinear Distance Vector (NDV).
% For each submanifold, the underlying geometric structures—including nodes, edges, adjacency matrices, and features—are coherently organized in the manifold space. 
% To alleviate the computational cost of Riemannian distance and preserve key geometric characteristics, we design a learnable distance memory bank to approximate manifold-based relationships with a novel Nonlinear Distance Vector (NDV).
% to  and maintaining predictive performance.
Finally, to jointly model the graphs constructed from both Riemannian and Euclidean representations, we design a new Fusion Graph Convolutional Network (FGCN) as the decision layer to jointly model the graphs constructed from both Riemannian and Euclidean representations. The learned features from the dual-path hybrid space are subsequently integrated through a weighted fusion mechanism, enhancing the predictive precision of the final output.
% This module effectively encodes ST dependencies within a unified geometric metric space, enabling more accurate forecasting. 

Our contributions are highlighted as follows:
\begin{itemize}
\item Submanifold-aware representation learning: We propose a novel architecture, SCS, which projects Euclidean features onto SPD submanifolds with overlapping temporal windows. This enables the model to explicitly capture dynamic geometric variations across ST dimensions.

\item Adaptive geometric correlation modeling: We design the ADB and introduce the NDV to learn fine-grained geometric correlations on the manifold. A memory-efficient distance bank is further developed to approximate Riemannian relationships while maintaining computational tractability.

\item Hybrid geometric decision layer: To jointly exploit Euclidean and Riemannian patterns, we design the new FGCN as the final decision layer. This component fuses dual-space features through a learnable weighted strategy, effectively capturing ST dependencies from a unified geometric perspective.

\item State-of-the-art performance: We conduct extensive experiments over the state-of-the-art (SOTA) models on three benchmark datasets. The results show that HSMGNN achieves improvements of up to 13. 8\% in the root mean squared error, demonstrating the effectiveness of our method in MTS forecasting.
\end{itemize}
%%%%%%%%%%%%%%%%%%%%%%%%%%%%%%%%%%%%%%%%%%%%%%%%%%%%%%%%%%%%%%%%%%%%%%%%
The remainder of this paper is organized as follows. Section~\ref{sec:related_work} reviews the existing literature related to MTS forecasting. Section~\ref{sec:problem_formulation} defines the problem and introduces necessary mathematical preliminaries. Section~\ref{sec:methodology} details our proposed HSMGNN, including the SCS, ADB, and FGCN. Section~\ref{sec:experiment} presents extensive experiments to validate the effectiveness of our method. Section~\ref{sec:conclusion} concludes the paper.

\section{Related Work}
\label{sec:related_work}
\subsection{MTS Forecasting in Euclidean Space}
Early studies on MTS forecasting primarily focused on modeling temporal dependencies in the Euclidean space. Classical statistical models, such as ARIMA~\cite{ARIMA}, leverage auto-regression, differencing, and moving average techniques to capture linear temporal relationships. Similarly, Gaussian processes~\cite{girard2002gaussian} model predictive distributions over time series using a probabilistic framework. However, these methods often struggle to handle nonlinearity, noise, and outliers in complex MTS data.

To better model non-linear patterns, DL-based methods have been widely adopted. Variants of Recurrent Neural Network (RNN), such as Long Short-Term Memory (LSTM)~\cite{hochreiter1997long,hua2019deep,malhotra2015long} and Gated Recurrent Unit (GRU)~\cite{cho2014learning,Tang_Yao_Sun_Aggarwal_Mitra_Wang_2020}, are frequently used for MTS forecasting. These models alleviate issues of RNNs like vanishing or exploding gradients and capture sequential dependencies effectively, albeit still within the Euclidean domain.

More recently, Transformer-based models~\citep{chen2024pathformer,jiang2022sequential,Li2023MTSMixersMT,wu2023timesnet,zhang2022crossformer} have achieved SOTA performance in long-range forecasting tasks due to their ability to model long-term dependencies via self-attention. However, most Transformer variants primarily emphasize temporal modeling, often overlooking the spatial dependencies inherent in MTS data. Moreover, the permutation-invariant nature of self-attention mechanisms may lead to temporal information loss~\cite{gruver2024large,Zeng2022AreTE}. In addition, these models generally rely heavily on timestamp-based positional encoding, limiting their generalization in scenarios with irregular or missing time intervals.
% \subsection{GNN methods for MTS data}

\subsection{MTS Forecasting in Non-Euclidean Space}
\subsubsection{GNN-based Methods}
As an alternative to the above methods, some studies~\cite{SAGDFN, FCSTGNN, zhao2019t} have adopted GNN, which is inherently suited for modeling complex spatial dependencies due to its capability of representing irregular structures and learning relationships beyond local neighborhoods. The work~\cite{zhao2019t} integrates Graph Convolutional Network (GCN) with GRU to capture ST dependencies. However, it relies on a fixed graph structure that cannot be dynamically updated. In SAGDFN~\cite{SAGDFN}, a slim adjacency matrix is utilized to capture overall spatial correlations at different time steps, but it overlooks the variations in correlations across different sensors. FCSTGNN~\cite{FCSTGNN} addresses this limitation by introducing a fully connected ST graph with a decay matrix, which explicitly models inter-sensor correlations across all timestamps. While the data structure is graph-based, the underlying graph topologies are often constructed using Euclidean metrics. These works often overlook the possibility that the data may naturally lie in a non-Euclidean space. This bias limits the expressiveness when dealing with complex geometric structures and evolving relationships among features of our interest.

\subsubsection{Manifold-based Methods}
Recent advances have begun to explore Riemannian or manifold-aware approaches, especially using the SPD manifold~\cite{li2021combination,pan2023isometric,Geometry_aware_2018} to capture the geometric characteristics of complex non-Euclidean data. While maintaining the data structure in the SPD manifold, researchers~\cite{SPDnet, Matt} transform the input SPD matrices to new compact SPD matrices. The study in \citet{Suh_Kim_2021_EEG_BANK} divides the whole non-linear data into several non-overlapping subspaces of the SPD manifold and learn the corresponding metric individually. This model shares the underlying representation and concatenates all partial solutions to obtain the final result under various metrics, leading to faster convergence speed and broader generality to non-stationary data. \citet{jeong2023DCMLTS} directly map the SPD manifold of MTS into the Cholesky space for easier computation and higher efficiency.
Despite their potential for superior geometric insights, SPD-based methods rely on costly Riemannian metric computations, limiting their real-world application to complex and non-linear MTS forecasting.

% To address the limitation of hybrid representation with computation constrain in existing approaches and comprehensively model ST dependencies within MTS data, we design \textit{HSMGNN}, a novel framework based on hybrid Euclidean-SPD Manifold representation for MTS data.
To overcome the limitations of hybrid representations in capturing geometric ST dependencies, especially the computational overhead introduced by operations on the SPD manifold, we propose \textit{HSMGNN}, a novel framework that combines Euclidean and SPD manifold features within a unified and efficient design.
% {\color{red} \textbf{COMMENT:} At the beginning of each subsection, please add a few sentences to summarise the commonality of the studies in the subsection, and clarify why they cannot solve the problem you try to solve. At the end of each section, please add a few sentences to highlight the common limitations that make them inapplicable to solving the problem of interest.}

%%%%%%%%%%%%%%%%%%%%%%%%%%%%%%%%%%%%%%%%%%%%%%%%%%%%%%%%%%%%%%%%%%%%%%%%
\section{Problem formulation}
\label{sec:problem_formulation}
In this section, we first give the mathematical definitions of MTS and multivariate network in our proposed method, then we formally present the MTS forecasting problem of interest.

\begin{definition}
\textbf{Multivariate Time Series}. The MTS records the quantities of interest generated by $N$ instances over $T$ time steps at a consistent interval, and each observation is a $ C$-dimensional variable, where $C$ is referred to as the channel. The observations made by all $N$ instances at time step $t$ are denoted by $\mathbf{M}_t\in \mathbb{R}^{N\times C}$.
\end{definition}

\begin{definition}\textbf{Multivariate Network.}
A multivariate network can be represented as a 
 time-index series of graph $\mathbb{G}= \left\{\mathbf{G}_1,\mathbf{G}_2, ..., \mathbf{G}_T \right\}$, where $\mathbf{G}_t=\left\{\mathbf{V}_t,\mathbf{E}_t,\mathbf{A}_t,\mathbf{U}_t \right\}$, $t=1,2,..., T$ denotes the static graph snapshot at timestamp $t$. 
 In the graph snapshot $\mathbf{G}_t$, $\mathbf{V}_t$ and $\mathbf{E}_t$ denote the node set and edge set, respectively. Each node corresponds to a sensor, while each edge indicates the dependency between sensor pairs. Through preprocessing and feature extraction of raw sensor data $\mathbf{M}_t$ at each timestep, we obtain the node feature matrix $\mathbf{U}_t \in \mathbb{R}^{N \times F}$, where $F$ specifies the feature dimension. The adjacency matrix $\mathbf{A}_t$ at time step $t$ encodes connectivity relationships, with its $(i,j)$-th element $A_{i,j}^t$ defined as
\begin{equation}
A_{i,j}^t = \begin{cases}
1, & \text{if } e_{i,j}^t \in \mathbf{E}_t \text{ connects nodes } i \text{ and } j \text{ at time } t ;\\
0, & \text{otherwise}.
\end{cases}
\end{equation}
Let the superscript $^{\rm s}$ indicate the SPD manifold space, and the superscript $^{\rm e}$ indicate the Euclidean space. The graph in the Euclidean space is denoted as $\mathbf{G}_t^{\rm e} = \{\mathbf{V}_t^{\rm e}, \mathbf{E}_t^{\rm e}, \mathbf{A}_t^{\rm e}, \mathbf{U}_t^{\rm e}\}$, where $\mathbf{V}_t^{\rm e} = \mathbf{V}_t$ is the node set, $\mathbf{A}_t^{\rm e} \in R^{N \times N}$ is the adjacency matrix preserving the Euclidean relationships, and $\mathbf{U}_t^{\rm e}$ captures the Euclidean features. The corresponding graph in the SPD manifold space is $\mathbf{G}_t^{\rm s} = \{\mathbf{V}_t^{\rm s}, \mathbf{E}_t^{\rm s}, \mathbf{A}_t^{\rm s}, \mathbf{U}_t^{\rm s}\}$, where $\mathbf{V}_t^{\rm s}$ encodes nodes as points on the SPD manifold, $\mathbf{A}_t^{\rm s}$ is the adjacency matrix preserving the SPD manifold relationships (e.g., via Log-Euclidean metrics), and $\mathbf{U}_t^{\rm s}$ captures the Riemannian features. 
% Thus, the graph in Euclidean space is denoted as $\mathbf{G}_t^{\rm e} = \{\mathbf{V}_t^{\rm e}, \mathbf{E}_t^{\rm e}, \mathbf{A}_t^{\rm e}, \mathbf{U}_t^{\rm e}\}$, where:

% \begin{itemize}
%     \item $\mathbf{V}_t^{\rm e} = \mathbf{V}_t$ is the node set,
%     \item $\mathbf{A}_t^{\rm e} \in R^{n \times n}$ is the adjacency matrix preserving Euclidean relationships.
%     \item $\mathbf{U}_t^{\rm e}$ captures Euclidean features.
% \end{itemize}

% The corresponding graph in the SPD manifold space is represented as $\mathbf{G}_t^{\rm s} = \{\mathbf{V}_t^{\rm s}, \mathbf{E}_t^{\rm s}, \mathbf{A}_t^{\rm s}, \mathbf{U}_t^{\rm s}\}$, where:

% \begin{itemize}
%     \item $\mathbf{V}_t^{\rm s}$ encodes nodes as points on the SPD manifold,
%     \item $\mathbf{A}_t^{\rm s}$ is the adjacency matrix preserving SPD manifold relationships (e.g., via Log-Euclidean metrics),
%     \item $\mathbf{U}_t^{\rm s}$ incorporate Riemannian features.
% \end{itemize}
\end{definition}

\begin{definition} \textbf{Multivariate Time Series Forecasting.}
% Let $\mathcal{D} = \left\{ (\mathbf{M}t, y_t) \right\}{t=1}^T$ denote the dataset, where $\mathbf{M}_t$ is the input matrix and $y_t$ is .
Given the labeled historical observations of MTS $\mathcal{D} =\left\{\mathbf{M}_{t},y_t\right\}_{t=1}^T$ in the past $T$ steps, the MTS forecasting problem of interest aims to learn a mapping $f_\theta$ to predict the corresponding label $y_t\in \mathbb{R}$ at time $t$:${f_\theta}\left(\mathbf{M}_{t}\right)=\hat{y}_{t}$,
where $\hat{y}_{t}$ is the predicted label of MTS at time step~$t$. 
\end{definition}
% We consider the original MTS data $\mathcal{M}\in \mathbb{R}^{N\times T\times D_{\rm{in}}}=\{\mathbf{x}_{t}\,|\,t = 1,...,T\}$ over $T$ time steps, where $\mathbf{x}_{t}$ is the data point at time step $t$, {\color{red} $N$ is the number of nodes}, and $D_{\rm{in}}$ is the input dimension. 
% Specifically, ST graph methods usually assume that the graph structure tends to be static all the time. However, in reality, data often have certain distribution deviations and dynamic changes over long time scales. This requires that the modeled graph also undergoes corresponding adaptive changes: the number of nodes, inherent features, edges, adjacency matrix, and underlying features evolve dynamically with changes. Our model is designed with hybrid stacking modules, and the time resolution (different division intervals for time series of the same length) and time distance (interaction with nodes of different distances within the same division interval) are considered. 
% In this paper, we wish to design a new method/model for MTS forecasting. A hybrid model is developed, where the input is the MTS data $\mathcal{M}$ and the output is the predicted value $\hat{y}_{t+T}$.
% More formally, the loss function L of the model can be described as follows.
% \begin{equation}
%     \begin{aligned}
%         \mathcal{L}(Y,\hat{Y};\mathcal{G}) &=  xxxxxxxx
%     \end{aligned}
% \end{equation}
In this paper, we propose HSMGNN to solve the problem of MTS forecasting on multivariate networks, where the key challenges include capturing complex ST dependencies, modeling hierarchical structural patterns, and preserving geometric constraints in the feature space. HSMGNN takes the multivariate time series $\mathcal{M}$ as input, along with a dynamic parallel graph in both Euclidean space $\mathbb{G}^{\rm e}= \left\{\mathbf{G}_1^{\rm e},\mathbf{G}_2^{\rm e},...,\mathbf{G}_T^{\rm e} \right\}$ and SPD manifold space $\mathbb{G}^{\rm s}= \left\{\mathbf{G}_1^{\rm s},\mathbf{G}_2^{\rm s}, ..., \mathbf{G}_T^{\rm s} \right\}$.
% $\mathbf{G}_t^{\rm e} = (\mathbf{V}_t, \mathbf{E}_t, \mathbf{A}_t^{\rm e}, \mathbf{U}_t^{\rm e})$  
The model then outputs the predicted label $\hat{y}_{t}$. The joint optimization objective combines both representations through the following loss function 
\begin{equation}
    \mathcal{L} = \frac{1}{T}\sum_{t=1}^{T}\|y_t - \hat{{y}}_t\|_2^2 .
\end{equation}
% where $d_{\mathcal{M}}$ is the Log-Euclidean distance on $SPD^d$, balancing forecasting accuracy in Euclidean space with geometric consistency on the manifold.

\section{Methodology}
\label{sec:methodology}
In this section, we first give the overall structure of HSMGNN for MTS forecasting. Then, we describe the detailed implementation of the SCS embedding, ADB layer, and FGCN in HSMGNN.
% {\color{red} \textbf{COMMENT:} Again, the problem solved by HSMGNN is still unclear in the above paragraph. MTS is not even mentioned here.}
\begin{figure*}[htb]
    \centering
    \includegraphics[width=0.96\textwidth]{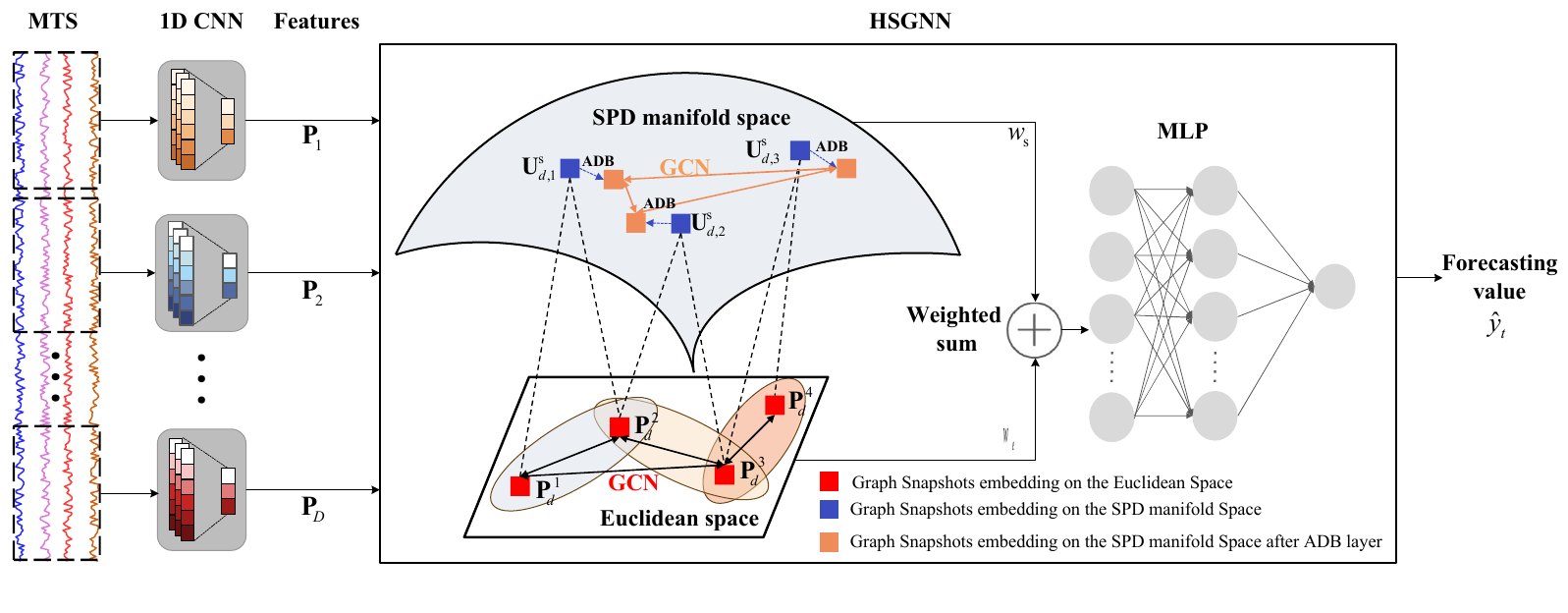} % Reduce the figure size so that it is slightly narrower than the column. Don't use precise values for figure width. This setup will avoid overfull boxes.
    
    \caption{Overall framework of the proposed HSMGNN. Given the input MTS, we segment it into patches using a sliding window of length $z_s = 2$ and stride $1$. 
Each patch is processed by a 1D CNN to extract Euclidean features $\mathbf{P}_1, \mathbf{P}_2, \dots, \mathbf{P}_D$. 
Taking the $d$-th dimension as an example, temporal features at four discrete time steps are denoted as $\mathbf{P}_d^1$ to $\mathbf{P}_d^4$. 
For each window, the two consecutive Euclidean points are projected onto the SPD manifold via their sample covariance, yielding $3$ overlapping Riemannian representations from $4$ Euclidean points. 
In the figure, red squares indicate Euclidean features, blue squares their SPD counterparts, and brown nodes features refined by the ADB layer. 
Graph convolution is applied independently in both spaces, and the results are fused via a weighted sum before being passed to a multilayer perceptron for prediction.
}
    \label{overall_structure}
    \end{figure*}
\subsection{Overview}
% We first elaborate on the general framework of our model. As illustrated in Fig.~\ref{overall_structure}, HSMGNN on the highest level consists of the Submanifold-Cross-Segment embedding, Adaptive-Distance-Bank, and the Fusion Graph Convolutional Network. To discover hidden geometric representation among MTS sensor nodes, the Submanifold-Cross-Segment embed the original data into Euclidean and SPD manifold space, which are later used as an input to the Fusion Graph Convolutional Network. To avoid the problem of highly computational cost of Riemannian distance, we introduce the Adaptive-Distance-Bank with trainable memory modules to capture ST dependencies under a determined time scale. Fig.~\ref{overall_structure} gives a demonstration of how the SCS and the ADB collaborate with each other.  To get the final prediction, the Fusion Graph Convolutional Network integrates the both hidden features in two spaces with MLP to the desired output dimension. In more detail, the core components of our model are illustrated in the following.

We begin by outlining the core design principles of the proposed HSMGNN framework, as depicted in Fig.~\ref{overall_structure}. At a high level, HSMGNN is composed of three major components: The Submanifold-Cross-Segment embedding module, the Adaptive-Distance-Bank, and the Fusion Graph Convolutional Network. The key idea behind HSMGNN is to jointly model MTS from both Euclidean and Riemannian geometric perspectives to capture diverse structural dependencies across sensors and times.

The SCS module encodes the raw MTS into two complementary geometric spaces, Euclidean and SPD manifolds, thereby revealing both linear and nonlinear ST correlations. These embeddings serve as the dual input to the subsequent fusion mechanism. 

The ADB module is a lightweight yet expressive memory-based module that learns adaptive temporal distances via a set of trainable kernels. It is designed to address the high computational cost associated with Riemannian geometry. This design significantly reduces computational overhead while preserving the ability to capture long-range dependencies under a pre-defined time granularity. 

As shown in Fig.~\ref{overall_structure}, the SCS and ADB modules extract geometry-aware representations, which are then integrated by the FGCN. The FGCN aggregates the features from both geometric views through a unified graph learning pipeline and projects them to the target prediction space via a Multi-Layer Perceptron (MLP). 

In the following subsections, we elaborate on the architecture and functionality of each of the components.
% {\color{red} \textbf{COMMENT:} Again, this above description does not show ``idea'', but just describe the steps.}

\subsection{Submanifold-Cross-Segment}
Given the input $\mathcal{D} \in \mathbb{R}^{N \times T \times C}$ with $C = 1$, we partition the temporal dimension $T$ into $L$ non-overlapping blocks using a sliding window of size and stride $W_{\mathrm{p}}$, where $L = \lfloor T / W_{\mathrm{p}} \rfloor$ and $\lfloor \cdot \rfloor$ denotes the floor operation. Each block has duration $W_{\mathrm{p}}$, forming the tensor $\mathbf{X} \in \mathbb{R}^{N \times W_{\mathrm{p}} \times L}$, defined as $\mathbf{X} = [\mathbf{X}_1, \mathbf{X}_2, \ldots, \mathbf{X}_L]$. The $l$-th block $\mathbf{X}_l \in \mathbb{R}^{N \times W_{\mathrm{p}}}$ covers the interval $[(l-1)W_{\mathrm{p}},\, lW_{\mathrm{p}})$, with $\mathbf{X}_l^t \in \mathbb{R}^N$ denoting the data at time step $t$ and $ 1 \leq l \leq L$:
\begin{equation}
\begin{aligned}
\mathbf{X}_l &= \left[ \mathbf{X}_l^{(l-1)W_{\mathrm{p}}},\, \ldots,\, \mathbf{X}_l^{lW_{\mathrm{p}} - 1} \right].
\end{aligned}
\label{eq:block_def}
\end{equation}

$\mathbf{X}$ is processed by a one-dimensional CNN to extract high-level features, yielding an output tensor $\mathbf{P} \in \mathbb{R}^{N \times W_{\mathrm{p}} \times D}$ given by
\begin{equation}
\mathbf{P} = \mathrm{ReLU}\left( \mathbf{W}_2^{\mathrm{p}} \ast \mathrm{ReLU}(\mathbf{W}_1^{\mathrm{p}} \ast \mathbf{X}) \right),
\label{eq:cnn_output}
\end{equation}
where $\ast$ denotes 1D convolution along the temporal axis; $\mathbf{W}_1^{\mathrm{p}}$ and $\mathbf{W}_2^{\mathrm{p}}$ are learnable convolutional kernels; $\mathrm{ReLU}(\cdot)$ is the rectified linear unit activation function; and $D$ is the number of output temporal blocks. The output $\mathbf{P}$ can be decomposed as $\mathbf{P} = [\mathbf{P}_1, \mathbf{P}_2, \ldots, \mathbf{P}_D]$, with each $\mathbf{P}_d \in \mathbb{R}^{N \times W_{\mathrm{p}}}$ representing the $d$-th $\left(1 \leq d \leq D\right)$ feature block.

Let $\delta \in (0, 1)$ denote the cross-decomposition ratio, which defines the sliding window length as $z_{\mathrm{s}} = \delta \cdot W_{\mathrm{p}}$ with a step size of 1. 
For each temporal block $\mathbf{P}_d \in \mathbb{R}^{N \times W_{\mathrm{p}}}$, we extract local segments $\mathbf{P}_d^{m:m + z_{\mathrm{s}}} \in \mathbb{R}^{N \times z_{\mathrm{s}}}$ over $M = W_{\mathrm{p}} - z_{\mathrm{s}} + 1$ consecutive windows, where $m = 1, \ldots, M$. 
Within each window, the sample covariance matrix is computed as:
\begin{equation}
\mathbf{U}_{d,m}^{\mathrm{s}} = \mathbf{P}_d^{m:m + z_{\mathrm{s}}} \left( \mathbf{P}_d^{m:m + z_{\mathrm{s}}} \right)^\top \in \mathbb{R}^{N \times N}.
\label{eq:covariance}
\end{equation}
These matrices are concatenated along the third dimension to form the tensor $\mathbf{U}_d^{\mathrm{s}} \in \mathbb{R}^{N \times N \times M}$, representing the SPD manifold features of the $d$-th block at scale $z_{\mathrm{s}}$.

Aggregating across all $D$ blocks yields the complete SPD feature tensor $\mathbf{U}^{\mathrm{s}} \in \mathbb{R}^{N \times N \times M \times D}$:
\begin{equation}
\mathbf{U}^{\mathrm{s}} = \left[ \mathbf{U}_1^{\mathrm{s}}, \mathbf{U}_2^{\mathrm{s}}, \ldots, \mathbf{U}_D^{\mathrm{s}} \right],
\label{eq:U_s}
\end{equation}
where each $\mathbf{U}_d^{\mathrm{s}} = \left[ \mathbf{U}_{d,1}^{\mathrm{s}}, \mathbf{U}_{d,2}^{\mathrm{s}}, \ldots, \mathbf{U}_{d,M}^{\mathrm{s}} \right]$ denotes the sequence of covariance matrices from block $d$.
% According to ~\citet{su2014submanifold}, the low-dimensional SPD manifold $\mathbf{U}^{\rm{s}}$ exhibits unique pattern characteristics with a common inherent long-term representation.

\subsection{Adaptive-Distance-Bank}
In the spatio-temporal graph, $\mathbf{G}_d^{\rm{s}} = \left\{\mathbf{V}_d^{\rm{s}}, \mathbf{A}_d^{\rm{s}}, \mathbf{E}_d^{\rm{s}}, \mathbf{U}_d^{\rm{s}} \right\}$ represents the $d$-th graph snapshot within the SPD manifold, where the patterns and intensities between nodes are typically characterized by the adjacency matrix. Conventionally, the adjacency matrix $\mathbf{A}_d^{\rm{s,o}} \in \mathbb{R}^{N \times N}$ is obtained directly through the dot product of the features $\mathbf{U}_d^{\rm{s}}$ with their own transpose:
\begin{equation}
{\bf{A}}_d^{{\rm{s}},{\rm{o}}} =  {\rm{softmax}}\left( {{\rm{ReLU}}\left( {{\bf{U}}_d^{\rm{s}}{{\left( {{\bf{U}}_d^{\rm{s}}} \right)}^{\rm{T}}}} \right)} \right),
\end{equation}
where $\mathrm{softmax}(\cdot)$ denotes the normalized exponential function, superscript $^{\rm{s},\rm{o}}$ denotes the original adjacency matrix defined on the SPD manifold, and $(\cdot)^{\text{T}}$ stands for transpose.

\begin{figure}[htb]
\centering
\includegraphics[width=0.98\columnwidth]{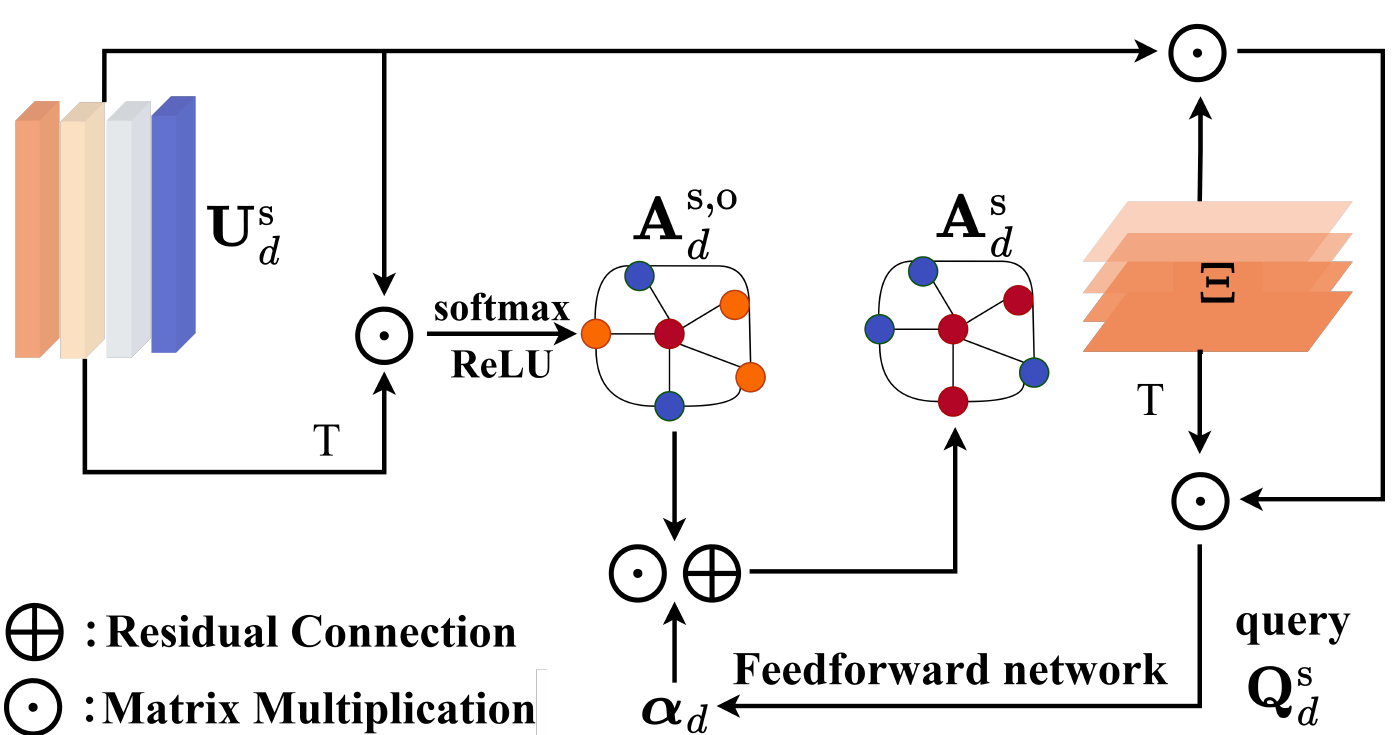} % Reduce the figure size so that it is slightly narrower than the column. Don't use precise values for figure width. This setup will avoid overfull boxes.
\caption{Adaptive Distance-Bank Layer. ${\mathbf{A}}_d^{\rm{s}}$ and ${\mathbf{A}}_d^{{\rm{s}},{\rm{o}}}$ capture different levels of connectivity in the graph, with node colors indicating different levels of structural hierarchy within the graph.}
\label{sub_aebb}
\end{figure}

As depicted in Fig.~\ref{sub_aebb}, the ADB layer learns the distribution and relative positions of multivariate sensor nodes in the SPD manifold space to generate a NDV $\boldsymbol{\alpha}_d \in \mathbb{R}^N$ that adjusts the adjacency matrix $\mathbf{A}_d^{\rm{s,o}}$. 
Given input $\mathbf{U}_d^{\rm{s}}$, we compute the bilinear mapping~\cite{SPDnet} to generate
\begin{equation}
\mathbf{Q}_d^{\rm{s}} = \Xi^\top \mathbf{U}_d^{\rm{s}} \Xi,
\end{equation}
{where $\mathbf{Q}_d^{\rm{s}} \in \mathbb{R}^{M_{\rm{q}} \times M_{\rm{q}} \times M}$ is the query matrix.} 

The trainable distance memory bank $\Xi \in \mathbb{R}^{N \times M_{\rm{q}}}$ is a matrix storing $N$ memory items, where $M_{\rm{q}}$ denotes the dimension of each memory item. This memory bank learns and stores typical patterns and features from historical data, and adaptively selects and combines memory items based on the SPD input characteristics to enable dynamic pattern adjustment.

The nonlinear distance vector $\boldsymbol{\alpha}_d$ is then extracted from $\mathbf{Q}_d^{\rm{s}}$ via a Feed Forward Network (FFN) $\Lambda(\cdot;\phi)$ with network parameter $\phi$ and hidden dimension $M_{\rm{d}}$:
\begin{equation}
\begin{aligned}
{{\boldsymbol{\alpha }}_d}  &=  \Lambda \left( {{{\left( \Xi  \right)}^{\rm{T}}}{\bf{U}}_d^{\rm{s}}\Xi ;\phi } \right)\\
& =  {\rm{cat}}\left( {\alpha _d^1,\alpha _d^2, ..., \alpha _d^N} \right),
\end{aligned}
\end{equation}
where $\mathrm{cat}(\cdot)$ denotes the concatenation operation. For a sensor node $i$, its corresponding nonlinear distance factor is $\alpha_d^i \in \mathbb{R}$, and $\alpha_d^i$ is the $i$-th scalar element of $\boldsymbol{\alpha}_d \in \mathbb{R}^N$ $\left(1 \leq i \leq N\right)$. 

Finally, the adjacency matrix $\mathbf{A}_d^{\rm{s}}$ for a temporal slice $\mathbf{U}_d^{\rm{s}}$ is obtained via residual connection:
\begin{equation}
{\mathbf{A}}_d^{\rm{s}} = {{\boldsymbol{\alpha }}_d}{\mathbf{A}}_d^{{\rm{s}},{\rm{o}}} + {\boldsymbol{A}}_d^{{\rm{s}},{\rm{o}}}.
\end{equation}
This mechanism enables dynamic optimization of both connection patterns and interaction intensities among nodes in $\mathbf{G}_d^{\rm{s}}$.

\subsection{Fusion GCN}
Using the SPD manifold feature tensor $\mathbf{U}^{\mathrm{s}} = \left[ \mathbf{U}_1^{\mathrm{s}}, \mathbf{U}_2^{\mathrm{s}}, \ldots, \mathbf{U}_D^{\mathrm{s}} \right]$ and the refined adjacency matrix sequence $\mathbf{A}^{\mathrm{s}} = \left[ \mathbf{A}_1^{\mathrm{s}}, \mathbf{A}_2^{\mathrm{s}}, \ldots, \mathbf{A}_D^{\mathrm{s}} \right]$, we perform graph convolution on the graph $\left\{\mathbf{G}_1^{\rm s},\mathbf{G}_2^{\rm s}, ..., \mathbf{G}_D^{\rm s} \right\}$ to capture the ST dependencies of the manifold features. 
Specifically, the graph convolution propagates and aggregates features as:
\begin{equation}
\begin{aligned}
{({\bf{U}}_d^{\rm{s}})_1} &  = {({\bf{A}}_d^{\rm{s}})^1}{\bf{U}}_d^{\rm{s}};\\
{({\bf{U}}_d^{\rm{s}})_2} &  = {({\bf{A}}_d^{\rm{s}})^2}{\bf{U}}_d^{\rm{s}};\\
 & ... \\
{({\bf{U}}_d^{\rm{s}})_{{r_{\rm{s}}}}} &  = {({\bf{A}}_d^{\rm{s}})^{{r_{\rm{s}}}}}{\bf{U}}_d^{\rm{s}},
\end{aligned}
\end{equation}
where $(\mathbf{A}_d^{\rm{s}})^{r_{\rm{s}}}$ denotes the adjacency matrix under the $r_{\rm{s}}$-th hop receptive field (considering only nodes within $r_{\rm{s}}$ hops), and the exponent $(\cdot)^{r_{\rm{s}}}$ represents the matrix power operation. 

In different receptive fields, we directly multiply the adjacency matrix by the node feature matrix $\mathbf{U}_d^{\rm{s}}$. The final node representation $\left(\mathbf{U}_d^{\rm{s}}\right)_{\text{final}}$ for graph snapshot $\mathbf{G}_d^{\rm{s}}$ is obtained by summing the results across all $r_{\rm{s}}$ receptive fields:
\begin{equation}
{\left( {{\bf{U}}_d^{\rm{s}}} \right)_{{\rm{final}}}} = {\rm{sum}}\left( {{{\left( {{\bf{U}}_d^{\rm{s}}} \right)}_1},{{\left( {{\bf{U}}_d^{\rm{s}}} \right)}_2}, ..., {{\left( {{\bf{U}}_d^{\rm{s}}} \right)}_{{r_s}}}} \right),
\end{equation}
where $\mathrm{sum}(\cdot)$ denotes the summation operation.
The complete state feature representation $\mathbf{U}^{\rm{s}}_{\rm{c}} \in \mathbb{R}^{D \times N \times F_{\rm{s}}}$ is then constructed by concatenating the final representations from all $D$ temporal blocks:
\begin{equation}
\mathbf{U}^{\rm{s}}_{\rm{c}} = {\rm{cat}}\left( {{{\left( {{\bf{U}}_1^{\rm{s}}} \right)}_{{\rm{final}}}},{{\left( {{\bf{U}}_1^{\rm{s}}} \right)}_{{\rm{final}}}}, \dots, {{\left( {{\bf{U}}_D^{\rm{s}}} \right)}_{{\rm{final}}}}} \right).
\end{equation}
Similarly, we perform analogous operations on the Euclidean graph $\left\{\mathbf{G}_1^{\rm e},\mathbf{G}_2^{\rm e}, ..., \mathbf{G}_K^{\rm e} \right\}$ ($K= D$) to obtain:
\begin{equation}
\begin{aligned}
{\left( {{\bf{U}}_k^{\rm{e}}} \right)_{{\rm{final}}}} &= {\rm{sum}}\left( {{{\left( {{\bf{U}}_k^{\rm{e}}} \right)}^1},{{\left( {{\bf{U}}_k^{\rm{e}}} \right)}^2}, ..., {{\left( {{\bf{U}}_k^{\rm{e}}} \right)}^{{r_{\rm{e}}}}}} \right);\\
\mathbf{U}^{\rm{e}}_{\rm{c}} &= {\rm{cat}}\left( {{{\left( {{\bf{U}}_1^{\rm{e}}} \right)}_{{\rm{final}}}},{{\left( {{\bf{U}}_2^{\rm{e}}} \right)}_{{\rm{final}}}}, ..., {{\left( {{\bf{U}}_K^{\rm{e}}} \right)}_{{\rm{final}}}}} \right).
\end{aligned}
\end{equation}
The final MTS prediction is obtained by processing both representations through an MLP:
\begin{equation}
    \begin{aligned}
{\bf{U}} &= {\rm{flatten}}\left( {{\rm{cat}}({w_{\rm{s}}}{\mathbf{U}^{\rm{s}}_{\rm{c}}},{w_{\rm{e}}}{\mathbf{U}^{\rm{e}}_{\rm{c}}})} \right);\\
{{\bf{H}}_1} &= \sigma \left( {{{\bf{W}}_1}{\bf{U}} + {{\bf{b}}_1}} \right);\\
{{\bf{H}}_2} &= \sigma \left( {{{\bf{W}}_2}{{\bf{H}}_1} + {{\bf{b}}_2}} \right);\\
{{\bf{H}}_3} &= \sigma \left( {{{\bf{W}}_3}{{\bf{H}}_2} + {{\bf{b}}_3}} \right);\\
\hat y &= {{\bf{W}}_4}{{\bf{H}}_3} + {{\bf{b}}_4},
\end{aligned}
\end{equation}
where $w_{\rm{s}}$ and $w_{\rm{e}}$ are the weights for $\mathbf{U}^{\rm{s}}_{\rm{c}}$ and $\mathbf{U}^{\rm{ce}}$, respectively; $\mathbf{W}_1 \in \mathbb{R}^{w_1 \times F_{\rm{u}}}$, $\mathbf{W}_2 \in \mathbb{R}^{w_2 \times w_1}$, $\mathbf{W}_3 \in \mathbb{R}^{w_3 \times w_2}$, and $\mathbf{W}_4 \in \mathbb{R}^{1 \times w_3}$ are the weight matrices of the MLP with $F_{\rm{u}} = K \times N \times F_{\rm{e}} + D \times N \times F_{\rm{s}}$; $\mathbf{b}_1 \in \mathbb{R}^{w_1}$, $\mathbf{b}_2 \in \mathbb{R}^{w_2}$, $\mathbf{b}_3 \in \mathbb{R}^{w_3}$, and $\mathbf{b}_4 \in \mathbb{R}$ are the bias vectors; $\mathrm{flatten}(\cdot)$ stands for the flattening operation.

\begin{table*}[htb]
\centering
\caption{Comparisons with SOTA in C-MAPSS (unit: cycle).}
\begin{tabular}{cc|cccccccccccc}
\hline
\multicolumn{2}{l|}{\multirow{2}{*}{Method}} & \multicolumn{3}{c}{FD001} & \multicolumn{3}{c}{FD002} & \multicolumn{3}{c}{FD003} & \multicolumn{3}{c}{FD004} \\ \cline{3-14} 
\multicolumn{2}{c|}{}                        & MAE   & MSE   & RMSE  & MAE   & MSE   & RMSE  & MAE   & MSE   & RMSE   & MAE   & MSE   & RMSE   \\ \hline
\multicolumn{2}{c|}{MegaCRN}                  & 12.70   & 314.04   & 17.72   & 16.23   & 454.42   & 21.37   & 8.09   & 148.31   & 12.17    & 13.17   & 337.77   & 18.37    \\
\multicolumn{2}{c|}{Crossformer}                   & 8.40   & 151.96   & 12.32   & 25.56   & 839.19   & 28.80   & 9.79   & 178.30   & 13.29    & 10.66   & 237.32   & 15.39    \\
\multicolumn{2}{c|}{SAGDFN}                  & \textbf{7.96}   & 149.26   & 12.22   & 10.33   & 217.29   & 14.74   & \textbf{6.92 }  & 123.43   & 11.10 & 10.52   & 248.75   & 15.77        \\
\multicolumn{2}{c|}{MAGNN}                   & 12.71   & 159.52   & 12.63   & 10.76   & 171.35   & 13.09   & 8.75   & 147.62   & 12.15    & 14.13   & 204.49   & 14.30    \\
\multicolumn{2}{c|}{FCSTGNN}                 & 9.16   & 135.02   & 11.62   & 10.16   & 170.04   & 13.04   & 8.16   & 132.71   &11.52   & 9.92   & 185.50  & 13.62   \\
\multicolumn{2}{c|}{MATT}                    & 11.25   & 210.25   & 14.50   & 19.38   & 571.10   & 23.90   & 15.00   & 351.56   & 18.75    & 16.13   & 436.48   & 20.90    \\
% \multicolumn{2}{l|}{GCN}                     & 888   & 888   & 888   & 888   & 888   & 888   & 888   & 888   & 888    & 888   & 888   & 888    
\hline
\multicolumn{2}{c|}{Ours}                    &{8.69 } & \textbf{132.94}   & \textbf{11.53}   & \textbf{10.05}  & \textbf{168.64}   & \textbf{12.98}   & 8.50   & \textbf{118.27}   & \textbf{10.88}    &\textbf{8.55}   & \textbf{165.73}   & \textbf{12.88}    \\ \hline
\end{tabular}

\label{tab:C-MAPSS-table}
\end{table*}

\subsection{Analysis of Computational Overhead}
% As mentioned before, the computation cost is pretty high at the SPD manifold,so xxxxxxx. The time complexity of operations in the SPD manifolds is dominated by the eigenvalue decomposition or Cholesky decomposition for the Log-Euclidean metric, with a computational complexity of ${\cal O}(E^3)$~\cite{jeong2023DCMLTS}, where $E$ denotes the embedding dimension of the matrix. As $E$ increases, the computational cost scales cubically. In contrast, the ADB achieves the equivalent geometric distance with a reduced complexity of ${\cal O}(E^2)$ for matrix multiplication. This quadratic scaling is computationally more efficient than the cubic scaling of conventional methods. Additionally, matrix multiplications with a complexity of ${\cal O}(N^2)$ are adopted to turn $\mathbf{A}_k^{\rm{s}}$ into influence towards $\mathbf{U}^{\rm{s}}$. As a result, the overall computational complexity for the equivalent Log-Euclidean metric in ADB is ${\cal O}(E^2+N^2)$, which is much lower than the conventional eigenvalue decomposition or Cholesky decomposition with no loss of performance to be studied in~\ref{sec:exp_res}.
As mentioned earlier, computing interdependencies between nodes on the SPD manifold typically involves expensive operations such as eigenvalue decomposition or Cholesky decomposition under the Log-Euclidean metric, with a time complexity of ${\cal O}(E^3)$~\cite{jeong2023DCMLTS}, where $E$ denotes the embedding dimension of the matrix. This cubic scaling significantly hinders scalability as $E$ grows. To address this issue, we introduce the ADB layer, an essential component of our framework, which takes advantage of a nonlinear network module (ADN) to approximate geometric relationships on the SPD manifold without explicit decomposition. Specifically, ADB achieves comparable interdependency modeling through matrix multiplications with a reduced complexity of ${\cal O}(E^2)$. Additional operations, such as transforming $\mathbf{A}_k^{\rm{s}}$ into influence over $\mathbf{U}^{\rm{s}}$, incur only a complexity of ${\cal O}(N^2)$. Consequently, the overall computational cost of the ADB module is ${\cal O}(E^2+N^2)$—lower than conventional Log-Euclidean approaches while retaining strong performance, as verified in Section~\ref{sec:exp_res}.
\section{Experiment}
\label{sec:experiment}
In this section, we evaluate HSMGNN against the SOTA methods on three benchmark datasets. We first assess the forecast performance of different methods in terms of commonly used metrics and then verify the efficacy of our proposed modules by an ablation study. Finally, we evaluate the impact of key parameters using a sensitivity analysis.
\subsection{Experiment Setting}
\textbf{Datasets.}
We examine our method in three different downstream tasks: RUL prediction, human activity recognition (HAR), and sleep stage classification (SSC). We used C-MAPSS~\cite{CMPASS_DATASET} for the prediction of RUL, UCI-HAR~\cite{HAR_anguita2012human} for HAR, and ISRUC-S3~\cite{ISRUC} for SSC. For all datasets, we adopt the same train-test split trick as~\cite{FCSTGNN}. 
% That is, for C-MAPSS, which includes four sub-datasets, we adopt the pre-defined train-test splits. The training dataset is further divided into 80\% and 20\% for training and validation. For HAR and ISRUC, we randomly split them into 60\%, 20\%, and 20\% for training, validating, and testing.

\noindent 
\textbf{Evalution.}
To evaluate the prediction performance of the RUL, we adopt the root mean square error (RMSE), the mean square error (MSE), and the mean absolute error (MAE), which are all widely used in MTS forecasting~\cite{chen2024biased,Megcan,SAGDFN,Matt,CL4ST}. Lower values of these indicators refer to better model performance. For the evaluation of HAR and SSC, we adopt accuracy (Accu) and macro-averaged F1 score (MF1) following previous studies~\cite{wang2024klinkllms,FCSTGNN}.

\noindent 
\textbf{Optimization Setting.}
All experiments are conducted with NVIDIA GeForce RTX 3090. We set the batch size as 32, choose ADAM as the optimizer with a learning rate of 1e-4 with early stopping, and train the model for 80 epochs. 
% More details can be found in our appendix. 

\noindent 
\textbf{Baseline Methods.}
We compare our model with the SOTA methods, which encompass SPD-based methods MATT~\cite{Matt}, Transformer-based Crossformer~\cite{zhang2022crossformer}, and GNN-based methods, including MAGNN~\cite{fu2020magnn}, MegaCRN~\cite{Megcan},SAGDFN~\cite{SAGDFN}, and FCSTGNN~\cite{FCSTGNN}. All methods are implemented following their original configurations.
% except GNN-based methods, where we replace their encoders with the same encoders used in our approach for a fair comparison.

\subsection{Comparisons with State-of-the-Art}
\label{sec:exp_res}
\begin{table}[htb]
\centering
\caption{Comparisons with SOTA in HAR (unit: \%).}
\begin{tabular}{cc|cccccccccccc}
\hline
\multicolumn{2}{c|}{\multirow{2}{*}{Method}} & \multicolumn{4}{c}{UCI-HAR} \\ \cline{3-6} 
\multicolumn{2}{c|}{}                        & \multicolumn{2}{c}{Accu}    & \multicolumn{2}{c}{MF1}      \\ \hline
\multicolumn{2}{c|}{MegaCRN}                   & \multicolumn{2}{c}{83.87}    & \multicolumn{2}{c}{83.39}       \\
\multicolumn{2}{c|}{Crossformer}              & \multicolumn{2}{c}{91.89}    & \multicolumn{2}{c}{91.84}       \\
\multicolumn{2}{c|}{SAGDFN}                   & \multicolumn{2}{c}{86.52}    & \multicolumn{2}{c}{74.66}      \\
\multicolumn{2}{c|}{MAGNN}               & \multicolumn{2}{c}{90.91}    & \multicolumn{2}{c}{90.79}     \\
\multicolumn{2}{c|}{FCSTGNN}              & \multicolumn{2}{c}{95.81}    & \multicolumn{2}{c}{95.82}    \\
\multicolumn{2}{c|}{MATT}                  & \multicolumn{2}{c}{93.93}    & \multicolumn{2}{c}{93.87}      \\ \hline
\multicolumn{2}{c|}{Ours}                  & \multicolumn{2}{c}{\textbf{96.33}}    & \multicolumn{2}{c}{\textbf{96.30}}      \\ \hline
\end{tabular}

\label{tab:HAR-table}
\end{table}

\begin{table}[htb]
\centering
\caption{Comparisons with SOTA in ISRUC-S3 (unit: \%).}
\begin{tabular}{cc|cccccccccccc}
\hline
\multicolumn{2}{c|}{\multirow{2}{*}{Method}} & \multicolumn{4}{c}{ISRUC-S3} \\ \cline{3-6} 
\multicolumn{2}{c|}{}                        & \multicolumn{2}{c}{Accu}    & \multicolumn{2}{c}{MF1}      \\ \hline
\multicolumn{2}{c|}{MegaCRN}                   & \multicolumn{2}{c}{76.74}    & \multicolumn{2}{c}{75.23}        \\
\multicolumn{2}{c|}{Crossformer}               & \multicolumn{2}{c}{60.82}    & \multicolumn{2}{c}{55.61}        \\
\multicolumn{2}{c|}{SAGDFN}                   & \multicolumn{2}{c}{29.20}    & \multicolumn{2}{c}{9.28}        \\
\multicolumn{2}{c|}{MAGNN}                & \multicolumn{2}{c}{68.13}    & \multicolumn{2}{c}{64.31}      \\
\multicolumn{2}{c|}{FCSTGNN}              & \multicolumn{2}{c}{{80.87}}    & \multicolumn{2}{c}{78.79}    \\
\multicolumn{2}{c|}{MATT}                   & \multicolumn{2}{c}{75.29}    & \multicolumn{2}{c}{71.00}       \\ \hline
\multicolumn{2}{c|}{Ours}                   & \multicolumn{2}{c}{\textbf{82.41}}    & \multicolumn{2}{c}{\textbf{81.25}}       \\ \hline
\end{tabular}

\label{tab:ISRUC-S3-table}
\end{table}
%ISRUC /home/fy/work/AAAI2025/Output/nni_experiments_ISRUC/0vm4k1hi/environments/local-env/trials/yGenp
Tables~\ref{tab:C-MAPSS-table}--\ref{tab:ISRUC-S3-table} strongly show that our model outperforms the SOTA methods in almost all cases. Among the SOTA models, FCSTGNN achieves relatively good performance for the sake of the Full-Connect-Graph Convolution backbone. Through Table~\ref{tab:C-MAPSS-table}, we can find that our model outperforms the SOTA methods in most metrics with improvements up to 13.8\% and 5.4\% over the second-best results regarding MAE and RMSE in FD004, respectively. Similar performance can be observed in UCI-HAR datasets in Table~\ref{tab:HAR-table}, where our method outperforms the second-best methods by 0.52\% regarding accuracy. In Table~\ref{tab:ISRUC-S3-table}, our model also has 1.54\% accuracy and 2.46\% MF1 gains. These advancements underline the necessity of fully capturing both Euclidean and Riemannian dependencies within MTS, thus enabling superior performance of HSMGNN.
%模块输入的复杂xxxx,输出的适合是yyyyy,y<<<<X+参考文献？？？？
\subsection{Ablation Study}
We conduct the ablation study to assess the effectiveness of our proposed modules in Tables~\ref{tab:C-MAPSS-Ablation-table}--\ref{tab:ISRUC_HAR-Ablation-table}. In the first variant `w/o SCS', we exclude the usage of the SCS. Instead, we follow the conventional methods~\cite{FCSTGNN} to separately construct and convolve graphs for each patch in the Euclidean space. The second variant `w/o ADB' involved incorporating the SCS but omitting the ADB, so the NDV is igored. Lastly, we introduce the third variant `w/o FGCN' by equipping the model with high-level features from the SPD manifold only. These variants are compared with the complete version. Tables~\ref{tab:C-MAPSS-Ablation-table} and \ref{tab:ISRUC_HAR-Ablation-table} present the ablation results across three datasets. 

Take the RMSE results on FD004 of C-MAPSS in Table~\ref{tab:C-MAPSS-Ablation-table} as an example. Comparing against the `w/o FGCN' variant, we observe that our complete method achieves a 9.87\% improvement, highlighting the necessity of capturing both Euclidean and Riemannian ST dependencies within MTS data. With the exclusion of ADB construction, there is a performance decrement of the `w/o ADB' variant, and the gap with the complete version is smaller than `w/o FGCN' , i.e., the performance gap is decreased to 7.07\%. This result indicates that FGCN effectively captures Riemannian spatiotemporal dependencies within MTS data by learning truly SPD manifold embeddings, even without explicit adjustment of the geometric distance. Finally, by incorporating the ADB and FGCN, we witness a further performance gap of the `w/o SCS' variant due to its lack of capturing both ST dependencies, narrowing the gap to 6.33\%. 

The above observations are consistent across UCI-HAR, ISRUC-S3, and other subdatasets of C-MAPSS. HSMGNN consistently outperforms most variants, validating the effectiveness of our proposed components. This comprehensive model leads to superior overall performance in various downstream tasks.

\begin{table*}[htb]
\centering
\caption{Ablation study in C-MAPSS (unit:cycle).}
\begin{tabular}{cc|cccccccccccc}
\hline
\multicolumn{2}{c|}{\multirow{2}{*}{Variants}} & \multicolumn{3}{c}{FD001} & \multicolumn{3}{c}{FD002} & \multicolumn{3}{c}{FD003} & \multicolumn{3}{c}{FD004} \\ \cline{3-14} 
\multicolumn{2}{c|}{}                        & MAE   & MSE   & RMSE  & MAE   & MSE   & RMSE  & MAE   & MSE   & RMSE   & MAE   & MSE   & RMSE   \\ \hline
\multicolumn{2}{c|}{w/o FGCN}                  & 9.61   & 163.44   & 12.79   & 16.93   & 621.82   & 24.94   & 9.24   & 167.81   & 12.95    & 10.53   & 204.29   & 14.29    \\ 
\multicolumn{2}{c|}{w/o SCS}                   & 9.16   & 160.47   & 12.63   & 12.63   & 265.63   & 16.32  & 10.20   & 184.14   & 13.57   & 9.28   & 191.30   & 13.75    \\
\multicolumn{2}{c|}{w/o ADB}                  & \textbf{7.76}   & 138.36   & 11.76   &\textbf{ 9.36 }  & \textbf{154.69}   & \textbf{12.44}   & \textbf{7.91}   & 133.40   & 11.55    & 9.88   & 192.10   & 13.86    \\
\hline
\multicolumn{2}{c|}{Complete}                    & 8.69   & \textbf{132.94 }  & \textbf{11.53}   & 10.05   & 168.64   & 12.98   & 8.50   & \textbf{118.27 }  & \textbf{10.88}    & \textbf{8.55 }  & \textbf{165.73}   & \textbf{12.88}    \\ \hline
\end{tabular}

\label{tab:C-MAPSS-Ablation-table}
\end{table*}

\begin{table}[htb]
\centering
\caption{Ablation study in ISRUC\&HAR (unit:\%).}
\begin{tabular}{cc|cccccccccccc}
\hline
\multicolumn{2}{c|}{\multirow{2}{*}{Method}} & \multicolumn{4}{c}{UCI-HAR} & \multicolumn{4}{c}{ISRUC-S3}\\ \cline{3-9} 
\multicolumn{2}{c|}{}                        & \multicolumn{2}{c}{Accu}    & \multicolumn{2}{c}{MF1a }   & \multicolumn{2}{c}{Accu}    & \multicolumn{2}{c}{MF1}   \\ \hline
\multicolumn{2}{c|}{w/o FGCN}                    & \multicolumn{2}{c}{95.07}    & \multicolumn{2}{c}{95.14}   & \multicolumn{2}{c}{78.59}    & \multicolumn{2}{c}{76.76}    \\
\multicolumn{2}{c|}{w/o SCS}                   & \multicolumn{2}{c}{93.95}    & \multicolumn{2}{c}{94.06} & \multicolumn{2}{c}{78.07}    & \multicolumn{2}{c}{76.56} \\
\multicolumn{2}{c|}{w/o ADB}               & \multicolumn{2}{c}{95.62}    & \multicolumn{2}{c}{95.67}  & \multicolumn{2}{c}{80.21}    & \multicolumn{2}{c}{78.55}  \\
\hline
\multicolumn{2}{c|}{Complete}                  & \multicolumn{2}{c}{\textbf{96.33}}    & \multicolumn{2}{c}{\textbf{96.30}}   & \multicolumn{2}{c}{\textbf{82.41}}    & \multicolumn{2}{c}{\textbf{81.25}}     \\ \hline
\end{tabular}

\label{tab:ISRUC_HAR-Ablation-table}
\end{table}%c}{80.79}    & \multicolumn{2}{c}{78.78}  \\
\subsection{Sensitivity Analysis}
In this subsection, we perform a sensitivity analysis on the SPD Cross-Decomposition Ratio, Distance-Bank Network Size, and Balance on FGCN. The key findings are presented below.
% Typical results are reported, and additional results can be found in our appendix.
\subsubsection{SPD Cross-Decomposition Ratio}
\begin{figure}[t]
\centering
\includegraphics[width=0.98\columnwidth]{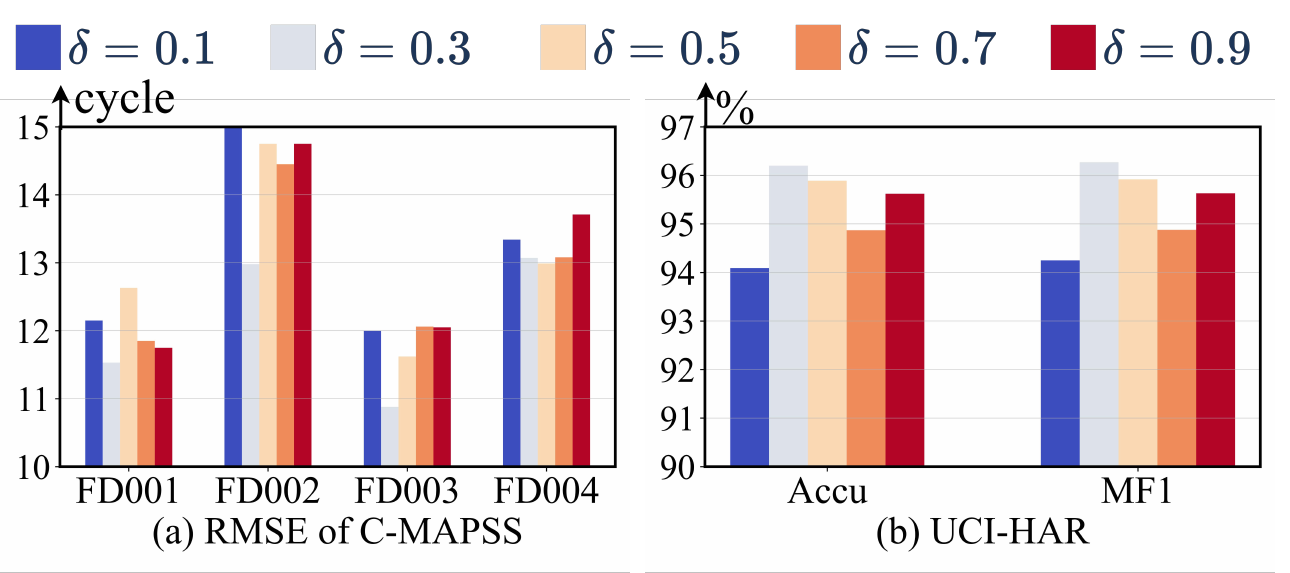} % Reduce the figure size so that it is slightly narrower than the column. Don't use precise values for figure width. This setup will avoid overfull boxes.
\caption{Sensitivity analysis for SPD cross-decomposition ratio $\delta$.}
\label{sub_sdcd_delta}
\end{figure}

We embed each MTS slice as cross-submanifolds for geometric graph construction, making the cross-decomposition ratio $\delta$ a parameter influencing the final graph construction. To evaluate its impact, we conduct the SPD cross-decomposition ratio analysis with $\delta \in \{0.1,0.3,0.5,0.7,0.9\}$, where different values of $\delta$ control the different cross lengths $z_{\rm{s}}$ of every single slice embedded in the SPD subspace. As shown in Fig.~\ref{sub_sdcd_delta}, it is observed that using small ratios results in poorer performance, highlighting the importance of embedding slices into a relatively long-term scale to effectively model global geometric ST dependencies. Moreover, performance gains diminish when a larger $\delta$ is introduced due to insufficient temporal representation. These results demonstrate that selecting an appropriate value for $\delta$ is crucial for optimal performance. For instance, in all cases, the model with $\delta=0.3$ outperforms its counterparts. In specific cases of UCI-HAR, introducing $\delta=0.5$ contributes to competitive performance compared to the best configuration.

\subsubsection{Adaptive-Distance-Bank Layer Size}
\begin{figure}[t]
\centering
\includegraphics[width=0.99\columnwidth]{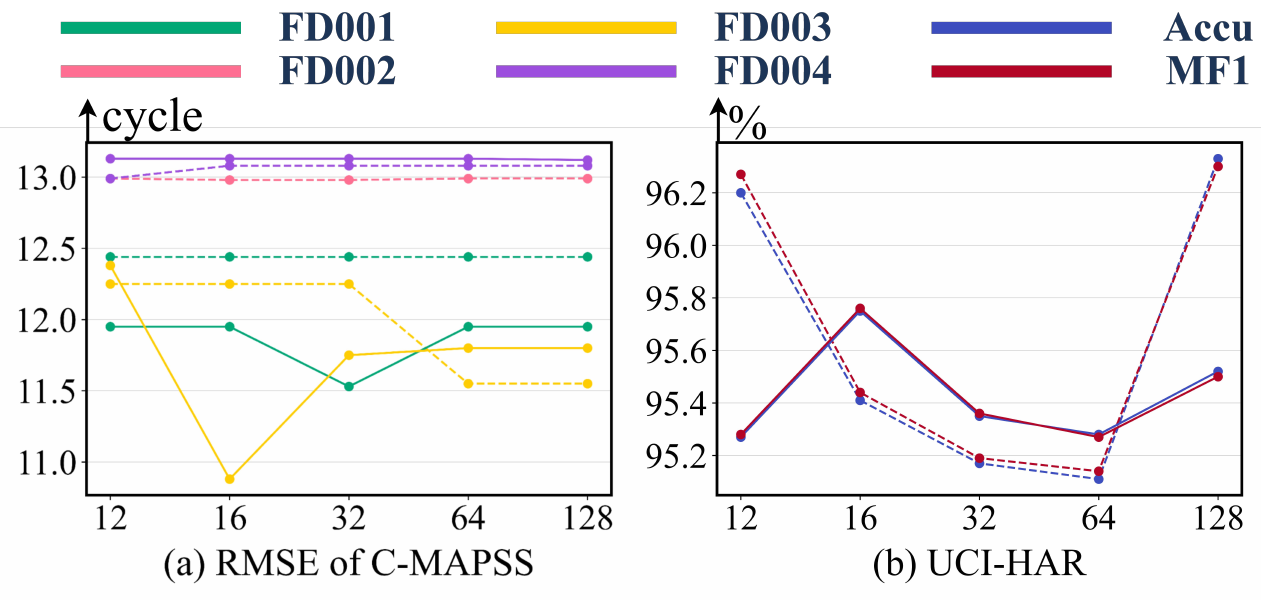} % Reduce the figure size so that it is slightly narrower than the column. Don't use precise values for figure width. This setup will avoid overfull boxes.
\caption{Sensitivity analysis for ADB layer size $M_{\rm{q}}$. The solid line is for $M_{\rm{q}}=64$ and the dashed line is for $M_{\rm{q}}=32$.}
\label{sub_sdcd_BANK}
\end{figure}

We design the ADB with specified sizes $M_{\rm{d}}$ and $M_{\rm{q}}$, capturing dynamic Riemannian ST dependencies within sensors of the MTS data. To assess the impact, we explore different FFN sizes $M_{\rm{d}} \in \{12, 16, 32, 64, 128\}$ and query sizes $M_{\rm{q}} \in \{32, 64\}$. Fig.~\ref{sub_sdcd_BANK} shows the results. For the C-MAPSS dataset, where the operational conditions are relatively complex (FD002 and FD004), performance remains pretty stable across varying network sizes. However, considering the RMSE of FD003, the optimal performance is observed at $M_{\rm{d}}=16$ and $M_{\rm{q}}=32$, with similar trends noted in FD001. Nevertheless, for datasets with longer time sequences, larger $M_{\rm{d}}$ values enhance performance. For instance, UCI-HAR achieves the optimal performance with $M_{\rm{d}}=128$ and $M_{\rm{q}}=32$, although overfitting occurs when the query size increases to 64. Meanwhile, competitive performance is observed with $M_{\rm{d}}=12$ and $M_{\rm{q}}=32$, demonstrating the generality of our model. These findings highlight the nuanced relationship between network size and performance, influenced by the specific dataset characteristics.

\subsubsection{Balance on FGCN}
\begin{figure}[t]
\centering
\includegraphics[width=0.98\columnwidth]{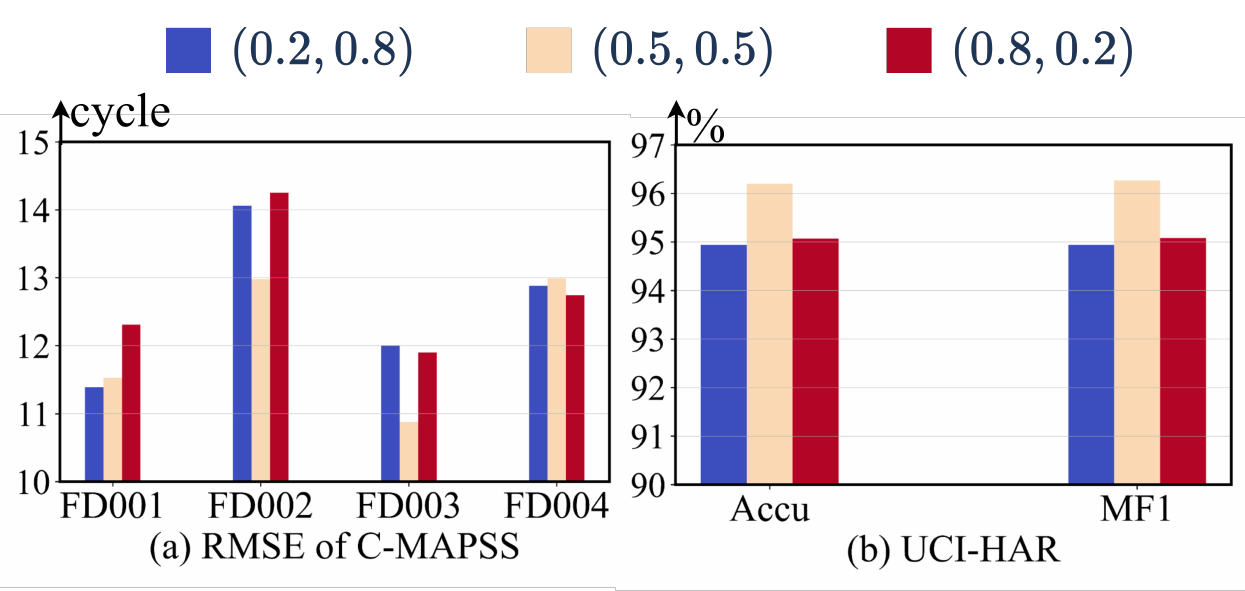} % Reduce the figure size so that it is slightly narrower than the column. Don't use precise values for figure width. This setup will avoid overfull boxes.
\caption{Sensitivity analysis for balance on FGCN.}
\label{sub_sdcd_HYBRID}
\end{figure}

Our approach leverages a compact SPD manifold embedding to capture cross-ST dependencies within MTS data from the Riemannian and Euclidean perspectives. To evaluate the impact of this hybrid geometric embedding, we evaluate the balance weight between the Riemannian and Euclidean components, using weight pairs $(0.2,0.8)$, $(0.5,0.5)$, and $(0.8,0.2)$. The former value in each pair represents the Riemannian weight, and the latter represents the Euclidean weight. Results are presented in Fig.~\ref{sub_sdcd_HYBRID}. The findings indicate that relying solely on Euclidean embeddings results in poorer performance, whereas the full model with hybrid embeddings yields enhanced results.
In Fig.~\ref{sub_sdcd_HYBRID}(a), which analyzes the C-MAPSS dataset using RMSE as the metric, a balanced weight ratio generally achieves better performance. For example, the weight configuration $(0.5,0.5)$ outperforms or closely matches other configurations in FD001, FD002, and FD003, demonstrating the value of integrating both geometric dependencies through our FGCN. However, in FD004, reducing the weight of the Riemannian component (from 0.8 to 0.5 or from 0.5 to 0.2) often decreases performance, likely because a larger Riemannian weight better captures SPD manifold features that Euclidean methods miss, which are crucial for accurately modeling dynamic ST dependencies. Similar trends are observed in UCI-HAR, where a balanced weight ratio of $(0.5,0.5)$ provides the best accuracy, as deviating from this balance leads to a loss of critical information.
Overall, these experiments suggest that the hybrid enhancement of the proposed FGCN is optimal for achieving the best performance.

\section{Conclusion}
\label{sec:conclusion}
This paper introduces a novel model, HSMGNN, for MTS forecasting with a hybrid
Euclidean-SPD manifold graph neural network. Our approach involves a Submanifold-Cross-Segment embedding, an Adaptive-Distance-Bank layer, and a Fusion Graph Convolutional Network. The SCS embeds different temporal dependencies into the smaller Riemannian patterns, and the ADB fine-tunes these patterns by introducing a novel nonlinear distance vector to facilitate dynamic ST graph learning on the SPD manifold space. Finally, FGCN is used to enhance the performance in both Euclidean and Riemannian spaces. Experimental results on three datasets show
its effectiveness over the previous state-of-the-art.
%%%%%%%%%%%%%%%%%%%%%%%%%%%%%%%%%%%%%%%%%%%%%%%%%%%%%%%%%%%%%%%%%%%%%%%%

%%% Use this environment to include acknowledgements (optional).
%%% This will be omitted in doubleblind mode.

% \begin{ack}
% This work was supported in part by the National Natural Science Foundation Program of China (NSFC) under Grants U21B2029, 62027805, and U21A20456, and in part by the Zhejiang Provincial Natural Science Foundation of China under Grant LR23F010006.
% \end{ack}

%%%%%%%%%%%%%%%%%%%%%%%%%%%%%%%%%%%%%%%%%%%%%%%%%%%%%%%%%%%%%%%%%%%%%%%%

%%% Use this command to include your bibliography file.

% \bibliography{m2356}

\end{document}